\documentclass[11pt]{article}

\usepackage[margin=1in]{geometry}
\usepackage{amsmath, amssymb}
\usepackage{graphicx}
\usepackage{booktabs}
\usepackage{tabularx}
\usepackage{array}
\usepackage{longtable}
\usepackage{pdflscape}
\newcolumntype{P}[1]{>{\raggedright\arraybackslash}p{#1}}
\usepackage{hyperref}
\usepackage{natbib}

\hypersetup{
  colorlinks=true,
  linkcolor=blue,
  citecolor=blue,
  urlcolor=blue
}

\title{Intra-tree Column Subsampling Hinders XGBoost Learning of Ratio-like Interactions}
\author{Mykola Pinchuk, PhD\thanks{Email: \texttt{pinchumkykola@gmail.com}. Companion repository (code and materials): \url{https://github.com/MykolaPinchuk/paper_xgb_colsample}.}\\Independent Researcher\\San Jose, USA}
\date{\today}

\begin{document}
\maketitle

\begin{abstract}
Many applied problems contain signal that becomes clear only after combining multiple raw measurements.
Ratios and rates are common examples.
In gradient boosted trees, this combination is not an explicit operation.
The model must synthesize it through coordinated splits on the component features.
This paper studies whether intra-tree column subsampling in XGBoost makes that synthesis harder.

Two synthetic data generating processes represent cancellation-style structure.
In both, two primitives share a strong nuisance factor, while the target depends on a smaller differential factor.
A log ratio of the primitives cancels the nuisance and isolates the signal.
Experiments vary \texttt{colsample\_bylevel} and \texttt{colsample\_bynode} over $s \in \{0.4, 0.6, 0.8, 0.9\}$, with emphasis on the mild range $s\ge 0.8$.
A control feature set includes the engineered ratio, which removes the need for synthesis.

Across both data generating processes, intra-tree column subsampling reduces test PR AUC in the primitives only setting. In the main data generating process such a relative decrease in PR AUC performance reaches 54\% when both hyperparameters are set to 0.4.
The effect largely disappears when the engineered ratio is present.
A path-based co-usage metric drops in the same cells where performance deteriorates.
These results support a practical recommendation.
If ratio-like structure is plausible, either avoid intra-tree column subsampling or include the intended ratio features.
\end{abstract}

\section{Introduction}

Tree-based models are a default choice in many applied machine learning settings.
XGBoost is widely used due to its accuracy and training efficiency \citep{chen2016xgboost}.
Practitioners often tune row subsampling and column subsampling to improve generalization.

Many real features are built from primitives.
Rates, ratios, and log ratios are common.
Advertising uses click through rate.
Operations uses defect rate.
Fraud modeling uses rates such as chargebacks per exposure.
In these cases, the primitives often share a large nuisance factor.
Exposure and scale are common examples.

If the model receives only primitives, it must combine them to cancel the nuisance.
If intra-tree column subsampling hides one primitive at a critical split, that combination can fail.
This paper focuses on this failure mode.

\begin{quote}
\noindent\textbf{Claim for practitioners.} Intra-tree column subsampling can hinder learning of important multi-feature interactions.
This paper establishes the effect for ratio-like and rate-like interactions that require combining multiple primitives to remove a shared nuisance factor.
\end{quote}

\section{Background}

\subsection{Column subsampling in XGBoost}

XGBoost supports three column subsampling parameters.
\texttt{colsample\_bytree} samples features once per tree.
\texttt{colsample\_bylevel} samples features once per depth level.
\texttt{colsample\_bynode} samples features independently at each node.

This paper studies \texttt{colsample\_bylevel} and \texttt{colsample\_bynode}.
Both change feature availability within a tree.
This matters when a good decision path needs coordinated access to multiple primitives.

A commonly used feature availability calculation can be misleading for trees.
At a single node, if a node samples a fraction $s$ of features, then a specific pair is both available with probability close to $s^2$ when the feature count is large.
A ratio-like interaction does not require both primitives to be available at the same node.
If a tree splits on $a$ at some depth, it only needs $b$ to be available at a descendant node.

Under \texttt{colsample\_bynode}, a feature is available at each node with probability close to $s$.
Along a path of length $L$, the probability that $b$ is available at least once is $1-(1-s)^L$.
For typical depths, this can be close to one even for $s=0.8$.
Therefore, a strict $s^2$ availability argument is too pessimistic.

The practical issue is greedy search under stochastic feature masking.
If the best split at a node uses a feature that is masked, the tree commits to a different partition.
This choice changes which subregions exist downstream.
It can remove opportunities for later splits to reconstruct the intended composite.
This sensitivity grows when the interaction is needed early in the tree, and when many distractor features compete for split gain.
This paper treats intra-tree column subsampling as search noise, not as a simultaneous availability constraint.

\subsection{Related work}

Randomization has a long history in tree ensembles as a way to reduce variance, decorrelate trees, and improve generalization.
Random forests sample features at each split \citep{breiman2001rf}, and extremely randomized trees introduce additional split randomization \citep{geurts2006extratrees}.
In gradient boosting, stochastic variants use subsampling to regularize stagewise fitting \citep{friedman2002stochastic}, and modern GBDT toolkits (XGBoost, LightGBM) expose both row and column subsampling controls for speed and generalization \citep{chen2016xgboost,ke2017lightgbm}.
Most of this literature emphasizes accuracy and efficiency, but gives little guidance on when within-tree feature masking can interfere with learning specific interaction structures.

Ratios, rates, and log-ratios are ubiquitous engineered features in applied modeling because they can normalize away shared exposure or scale effects.
In compositional data analysis, log-ratio transforms are a standard tool for removing common-scale degrees of freedom \citep{aitchison1982compositional}.
For tree models, interactions are represented implicitly via axis-aligned partitions. Expressing a composite such as a ratio from raw primitives may require coordinated splits and sufficient depth \citep{hastie2009esl}.
This paper connects these themes by isolating a cancellation-style setting where the intended signal is concentrated in a (log-)ratio.
It quantifies how intra-tree column subsampling can add search noise that disrupts that coordination.

This paper studies cancellation-style interactions.
Two primitives share a strong nuisance factor $U$.
They carry opposite signed signal $V$.
A log ratio cancels $U$.

A canonical continuous construction is:
\begin{align}
\log a &= U + V + \epsilon_a, \\
\log b &= U - V + \epsilon_b, \\
r &= \log a - \log b \approx 2V + (\epsilon_a - \epsilon_b).
\end{align}

The target depends on $V$, not on $U$.
The model receives either $(a,b)$ or $(a,b,r)$.
When $r$ is included, the intended interaction is provided.

\section{Experimental setup}

\subsection{Feature sets}

Two feature sets isolate the role of synthesis.
\begin{itemize}
\item \texttt{F0}: primitives only.
\item \texttt{F1}: primitives plus the engineered ratio feature.
\end{itemize}

In both cases, the model also receives noise features.
These distractors are correlated with the nuisance factor.
In all experiments, there are two signal primitives and $p_{\mathrm{noise}}=120$ distractors.
In F1, the engineered ratio feature adds one more feature.
Therefore, the feature counts are 122 for F0 and 123 for F1.

The datasets use train=25{,}000, validation=10{,}000, and test=10{,}000 examples.
The test prevalence is fixed at $\pi=0.05$. Thus, the results in this paper generalize to the imbalanced classification problems common in practice.
A random ranking has expected PR-AUC equal to $\pi$.
This paper reports absolute PR-AUC deltas, and also reports relative deltas in the Appendix.

\subsection{Column subsampling arms}

A baseline configuration uses no intra-tree column subsampling.
Arms apply within-tree subsampling with $s \in \{0.4, 0.6, 0.8, 0.9\}$.
\begin{itemize}
\item \texttt{C0}: \texttt{colsample\_bylevel=1.0}, \texttt{colsample\_bynode=1.0}.
\item \texttt{C1}: \texttt{colsample\_bylevel=$s$}, \texttt{colsample\_bynode=1.0}.
\item \texttt{C2}: \texttt{colsample\_bylevel=1.0}, \texttt{colsample\_bynode=$s$}.
\item \texttt{C3}: \texttt{colsample\_bylevel=$s$}, \texttt{colsample\_bynode=$s$}.
\end{itemize}

\subsection{Data generating processes}

Two data generating processes are used.

\noindent DGP-A (continuous log ratio).
Latents $U$ and $V$ generate primitives $a$ and $b$ as in the equations above.
The engineered control feature is $r=\log a - \log b$.

\noindent DGP-B (count plus exposure).
An exposure latent $E$ is shared by two count variables.
A differential latent $V$ changes their rates in opposite directions.
The engineered control feature is a smoothed log rate difference, written as $\log(A+s_0) - \log(B+s_0)$.
The smoothing constant $s_0$ avoids issues at low counts.

\subsection{Metrics}

The primary metric is test PR AUC.
Secondary metrics (including ROC AUC) support interpretation.
All results are paired deltas against a matched baseline with no intra-tree column subsampling (C0, $s=1.0$).
The paper reports the mean paired delta across replicated runs.
Uncertainty intervals are normal approximation 95\% confidence intervals on this mean.
Appendix \ref{app:ci} gives the exact formula.

\noindent Path co-usage.
\texttt{cooc\_path\_mean} measures how often both primitives appear on the same root to leaf path.
Paths are weighted by cover.
A model that relies on the cancellation interaction should use both primitives along important paths.

\noindent Latent alignment.
\texttt{latent\_corr} is the correlation between the model score and the true latent signal $V$ on the test set.
This is available because the data are synthetic.

\section{Results}

\begin{figure}[t]
\centering
\includegraphics[width=\textwidth]{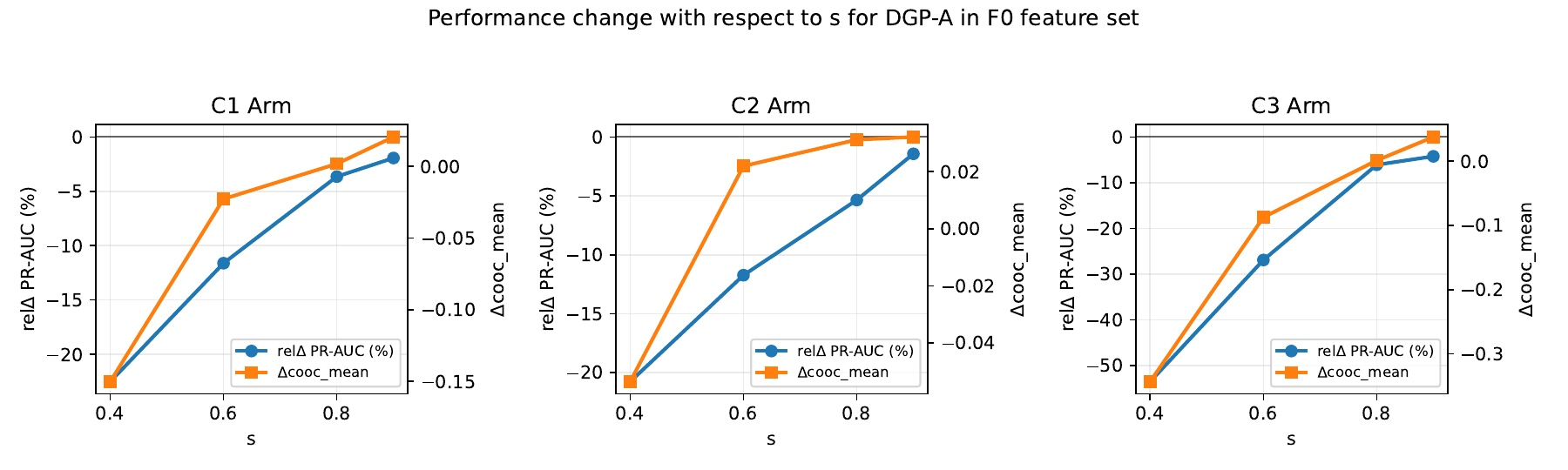}
\caption{DGP-A (primitives only, F0). Relative change in test PR-AUC (left axis) and change in path co-usage (right axis) as a function of the within-tree sampling rate $s$ for arms C1--C3.}
\label{fig:dgpA_main}
\end{figure}

Figure \ref{fig:dgpA_main} summarizes DGP-A in the primitives-only setting (F0).
As the within-tree sampling rate $s$ decreases, test PR-AUC declines and the path co-usage metric shifts in tandem.
The degradation is the largest when both \texttt{colsample\_bylevel} and \texttt{colsample\_bynode} are active (C3), consistent with subsampling disrupting coordinated use of the two primitives.

Table \ref{tab:main_dgpA_c3} focuses on the strongest subsampling arm (C3) and contrasts F0 with the engineered-feature control (F1).
In F0, C3 reduces PR-AUC by 0.010 at $s=0.8$ and 0.007 at $s=0.9$ (6.1\% and 4.2\% relative to the corresponding baselines).
When $s$ is pushed lower, the performance drop grows rapidly (down to $\Delta$PR-AUC=-0.090 at $s=0.4$), which corresponds to a 53.5\% relative decrease versus baseline.
In F1, PR-AUC deltas under C3 are small in magnitude, indicating that providing the engineered ratio largely removes the sensitivity.

\begin{table}[t]
\centering
\scriptsize
\setlength{\tabcolsep}{3.0pt}
\renewcommand{\arraystretch}{1.15}
\caption{DGP-A (continuous log ratio), arm C3. Panels A and B report paired deltas versus the baseline C0. Panel C reports baseline raw metric values under C0 ($s=1.0$). Relative deltas are percent change relative to the corresponding baseline.}
\label{tab:main_dgpA_c3}
\vspace{4pt}

\noindent\textbf{Panel A: F0 (primitives only)}\\[-2pt]
\resizebox{\textwidth}{!}{%
\begin{tabular}{llllllrrr}
\toprule
Arm & $s$ & \shortstack{$\Delta$PR-AUC\\(95\% CI)} & \shortstack{rel$\Delta$PR-AUC\\(\%)} & \shortstack{$\Delta$ROC-AUC\\(95\% CI)} & \shortstack{rel$\Delta$ROC-AUC\\(\%)} & $\Delta$cooc\_path\_mean & $\Delta$latent\_corr & $n$ \\
\midrule
C3 & 0.4 & -0.0897 [-0.0960, -0.0833] & -53.5 & -0.1705 [-0.1880, -0.1530] & -22.6 & -0.343 & -0.469 & 12 \\
C3 & 0.6 & -0.0459 [-0.0577, -0.0341] & -26.9 & -0.0693 [-0.0877, -0.0509] & -9.2 & -0.087 & -0.207 & 12 \\
C3 & 0.8 & -0.0104 [-0.0172, -0.0036] & -6.1 & -0.0184 [-0.0280, -0.0089] & -2.4 & 0.001 & -0.041 & 12 \\
C3 & 0.9 & -0.0071 [-0.0116, -0.0026] & -4.2 & -0.0095 [-0.0153, -0.0037] & -1.3 & 0.038 & -0.032 & 12 \\
\bottomrule
\end{tabular}%
}

\vspace{6pt}
\noindent\textbf{Panel B: F1 (primitives + engineered ratio)}\\[-2pt]
\resizebox{\textwidth}{!}{%
\begin{tabular}{llllllrrr}
\toprule
Arm & $s$ & \shortstack{$\Delta$PR-AUC\\(95\% CI)} & \shortstack{rel$\Delta$PR-AUC\\(\%)} & \shortstack{$\Delta$ROC-AUC\\(95\% CI)} & \shortstack{rel$\Delta$ROC-AUC\\(\%)} & $\Delta$cooc\_path\_mean & $\Delta$latent\_corr & $n$ \\
\midrule
C3 & 0.4 & -0.0085 [-0.0123, -0.0048] & -3.5 & -0.0118 [-0.0187, -0.0050] & -1.5 & 0.064 & 0.020 & 12 \\
C3 & 0.6 & -0.0020 [-0.0076, 0.0036] & -0.8 & -0.0062 [-0.0111, -0.0013] & -0.8 & 0.086 & -0.003 & 12 \\
C3 & 0.8 & 0.0027 [-0.0002, 0.0057] & 1.2 & -0.0019 [-0.0042, 0.0004] & -0.2 & 0.045 & -0.011 & 12 \\
C3 & 0.9 & 0.0013 [-0.0010, 0.0036] & 0.6 & -0.0008 [-0.0028, 0.0012] & -0.1 & 0.016 & 0.005 & 12 \\
\bottomrule
\end{tabular}%
}

\vspace{6pt}
\noindent\textbf{Panel C: Baseline C0 (raw values, $s=1.0$)}\\[-2pt]
\begin{tabular}{lrrrrr}
\toprule
Feature set & PR-AUC & ROC-AUC & latent\_corr & cooc\_path\_mean & $n$ \\
\midrule
F0 & 0.1679 & 0.7544 & 0.7074 & 0.5596 & 12 \\
F1 & 0.2381 & 0.8154 & 0.8146 & 0.0000 & 12 \\
\bottomrule
\end{tabular}%
\end{table}

\begin{figure}[t]
\centering
\includegraphics[width=\textwidth]{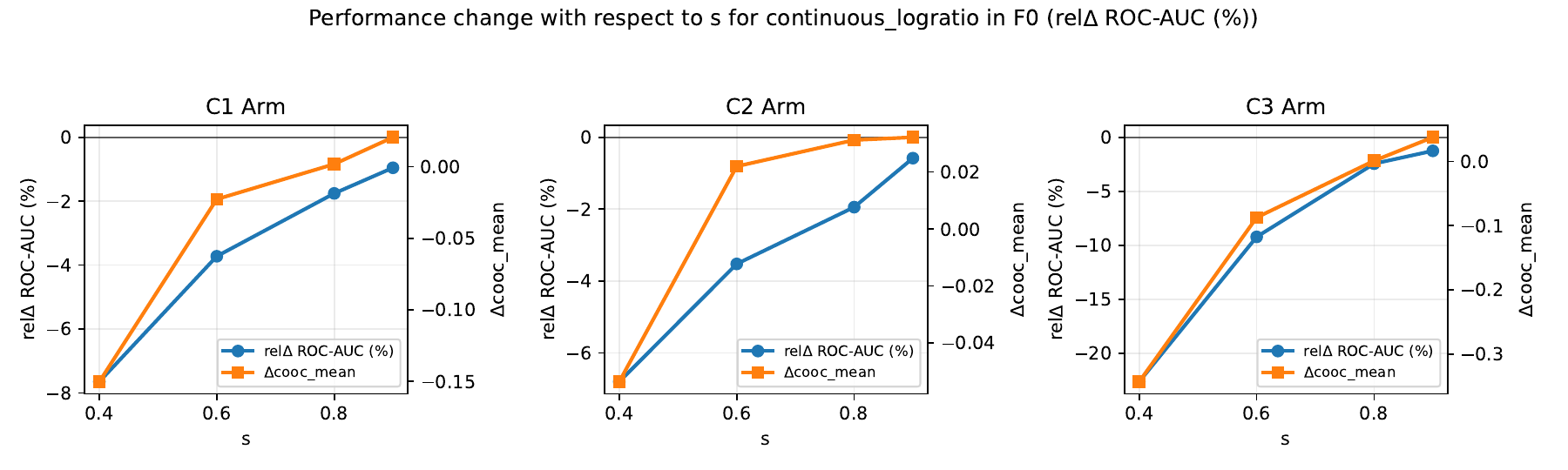}
\caption{DGP-A (F0). Relative $\Delta$ROC-AUC (left axis) and $\Delta$\texttt{cooc\_path\_mean} (right axis) across $s$ and arms C1--C3.}
\label{fig:dgpA_rel_roc}
\end{figure}

Figure \ref{fig:dgpA_rel_roc} provides a more complete sweep over $s$ and arms C1--C3.
Appendix Figures \ref{fig:low_s_A}--\ref{fig:boundary} provide additional sweeps and diagnostics.
Appendix Figure \ref{fig:app_dgpB_rel_pr} shows that the same qualitative pattern holds in DGP-B, with the strongest degradation under C3 at aggressive subsampling.

\section{Discussion}

\subsection{What these results do and do not say}

These experiments show a specific fragility.
When the true signal is accessible mainly through a ratio-like cancellation interaction, intra-tree column subsampling can significantly reduce performance.
The effect is visible even at mild settings.
The effect largely disappears when the engineered ratio is present in the feature set.

These results do not imply that intra-tree subsampling is always harmful.
It can still reduce overfitting in other settings.
This paper focuses on one interaction geometry that is common in practice.

\subsection{Practical guidance}

Practitioners often use column subsampling for speed and regularization.
The results suggest simple safeguards.

\begin{itemize}
\item If a ratio or rate feature is important, include it explicitly.
Do not rely on the model to always synthesize it from primitives.
\item If engineered ratios are not available, start with \texttt{colsample\_\allowbreak bylevel=1.0} and \texttt{colsample\_\allowbreak bynode=1.0}.
Tune other regularizers first.
\item If intra-tree subsampling is needed, prefer mild values and validate carefully.
Sensitivity is task dependent.
\end{itemize}

\subsection{Limitations and next steps}

DGP-B appears to exhibit significant degradation in performance only at aggressive values of s. Future work should focus on better understanding heterogeneity in results across DGP-A and DGP-B.

Effect sizes depend strongly on the subsampling rate.
At mild subsampling ($s=0.8$--$0.9$), the largest PR-AUC drop in these results is about 0.010.
When subsampling is more aggressive ($s=0.4$), the drop can reach about 0.090.
This is not a large change in isolation.
In many imbalanced classification applications, PR-AUC improvements of a few thousandths are operationally meaningful.
In other settings, this magnitude can fall below decision noise.
Therefore, the safest recommendation is empirical.
If ratio-like structure is plausible, validate intra-tree subsampling choices against a baseline with $s=1.0$.

Feature dimensionality matters for this mechanism.
These experiments use 120 distractors.
The effect is expected to strengthen when more distractors compete for split gain, or when the ratio needs to be expressed at shallow depths.
This paper does not vary feature count.
It treats this as an important extension.

This draft studies two synthetic data generating processes.
It uses mild subsampling levels.
It uses a single model capacity per setting.
A broader study could vary feature counts, depth, interaction order, and distributional shifts.
It could also add real world datasets where ratios are known to matter.

\section{Conclusion}

Intra-tree column subsampling in XGBoost can hinder learning of ratio-like interactions from primitives.
The effect appears in two cancellation-style synthetic tasks.
Providing the engineered ratio largely removes the sensitivity.
For applied work, this supports a conservative default.
Either include key ratio features or avoid intra-tree subsampling when such structure is plausible.

\clearpage
\appendix

\section*{Appendix}
This appendix provides additional sweeps, tables, and supporting definitions.

\section{Comprehensive sweep figures and tables}
\label{app:comp}

\begin{figure}[!ht]
\centering
\includegraphics[width=0.96\textwidth]{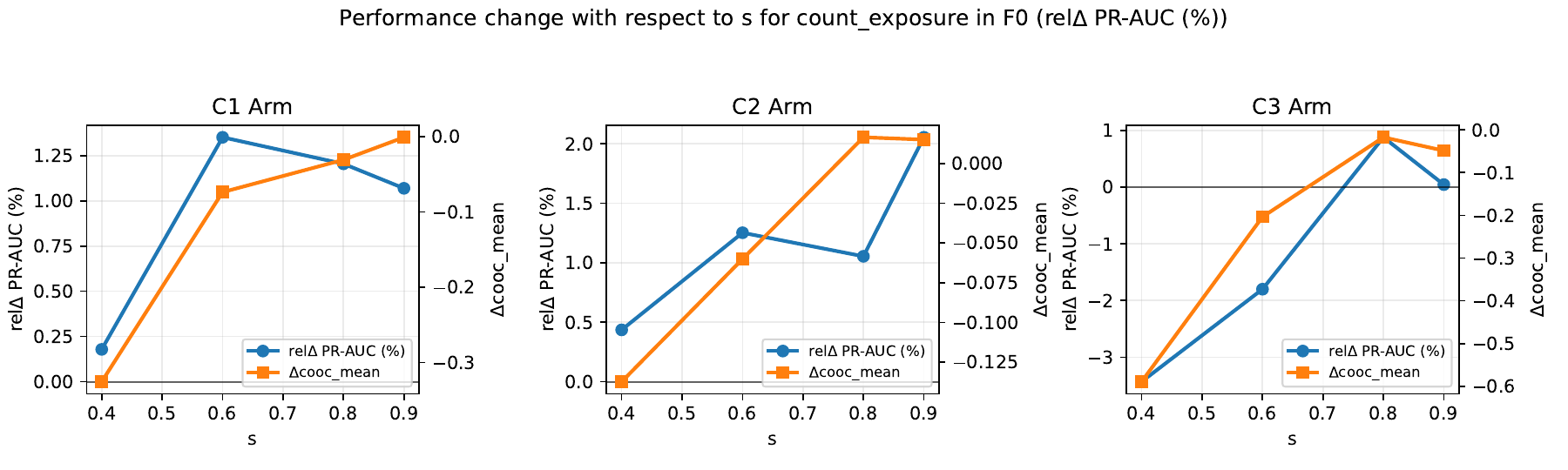}
\caption{DGP-B (F0). Relative $\Delta$PR-AUC (left axis) and $\Delta$\texttt{cooc\_path\_mean} (right axis) across $s$ and arms C1--C3.}
\label{fig:app_dgpB_rel_pr}
\end{figure}

\begin{figure}[!ht]
\centering
\includegraphics[width=0.96\textwidth]{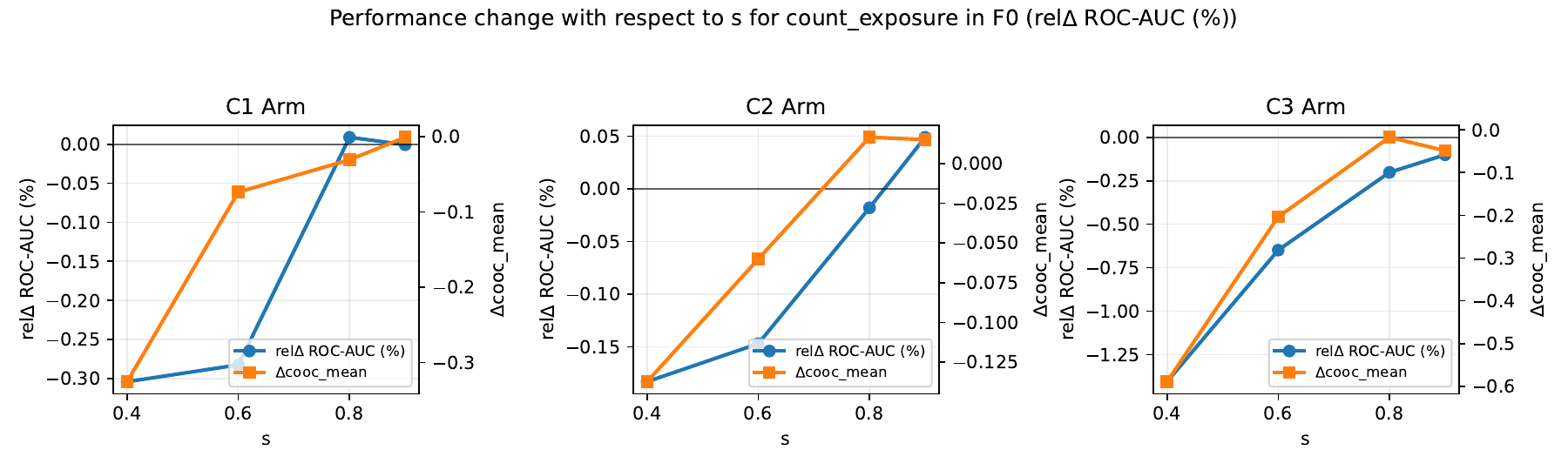}
\caption{DGP-B (F0). Relative $\Delta$ROC-AUC (left axis) and $\Delta$\texttt{cooc\_path\_mean} (right axis) across $s$ and arms C1--C3.}
\label{fig:app_dgpB_rel_roc}
\end{figure}

\begin{table}[t]
\centering
\small
\setlength{\tabcolsep}{4pt}
\renewcommand{\arraystretch}{1.1}
\begin{tabular}{llrrrrr}
\toprule
DGP & Feature set & PR-AUC & ROC-AUC & \texttt{latent\_corr} & \texttt{cooc\_mean} & $n$ \\
\midrule
A & F0 & 0.1679 & 0.7544 & 0.7074 & 0.5596 & 12 \\
A & F1 & 0.2381 & 0.8154 & 0.8146 & 0.0000 & 12 \\
B & F0 & 0.1949 & 0.7872 & 0.7844 & 0.9551 & 12 \\
B & F1 & 0.1970 & 0.7907 & 0.7830 & 0.0206 & 12 \\
\bottomrule
\end{tabular}
\caption{Baseline (C0, $s=1.0$) test metrics used to compute relative deltas in Tables~\ref{tab:app_v3_f0}--\ref{tab:app_v3_f1}. DGP A is continuous log ratio; DGP B is count plus exposure.}
\label{tab:app_v3_baseline}
\end{table}

\begin{landscape}
\begin{center}
\scriptsize
\setlength{\tabcolsep}{3pt}
\renewcommand{\arraystretch}{1.05}
\begin{longtable}{llr P{4.2cm} r P{4.2cm} r r r r}
\caption{Full $s$ sweep results for feature set F0 (primitives only). DGP A is continuous log ratio; DGP B is count plus exposure. Deltas are paired against the matched baseline (C0, $s=1.0$). Relative deltas are percent change relative to the corresponding baseline metric.}\label{tab:app_v3_f0}\\
\toprule
DGP & Arm & $s$ & $\Delta$PR-AUC (95\% CI) & rel$\Delta$PR (\%) & $\Delta$ROC-AUC (95\% CI) & rel$\Delta$ROC (\%) & $\Delta$\texttt{cooc\_mean} & $\Delta$\texttt{latent\_corr} & $n$ \\
\midrule
\endfirsthead
\multicolumn{10}{l}{\textit{Table \thetable\ continued from previous page}}\\
\toprule
DGP & Arm & $s$ & $\Delta$PR-AUC (95\% CI) & rel$\Delta$PR (\%) & $\Delta$ROC-AUC (95\% CI) & rel$\Delta$ROC (\%) & $\Delta$\texttt{cooc\_mean} & $\Delta$\texttt{latent\_corr} & $n$ \\
\midrule
\endhead
\bottomrule
\endfoot
\bottomrule
\endlastfoot
A & C1 & 0.4 & -0.0381 [-0.0458, -0.0304] & -22.5 & -0.0576 [-0.0736, -0.0416] & -7.6 & -0.150 & -0.158 & 12 \\
A & C1 & 0.6 & -0.0198 [-0.0253, -0.0144] & -11.6 & -0.0282 [-0.0413, -0.0151] & -3.7 & -0.023 & -0.082 & 12 \\
A & C1 & 0.8 & -0.0063 [-0.0121, -0.0006] & -3.6 & -0.0134 [-0.0247, -0.0022] & -1.8 & 0.002 & -0.030 & 12 \\
A & C1 & 0.9 & -0.0032 [-0.0082, 0.0018] & -1.9 & -0.0072 [-0.0148, 0.0004] & -1.0 & 0.020 & -0.019 & 12 \\
A & C2 & 0.4 & -0.0359 [-0.0466, -0.0251] & -20.8 & -0.0516 [-0.0653, -0.0379] & -6.8 & -0.054 & -0.143 & 12 \\
A & C2 & 0.6 & -0.0202 [-0.0293, -0.0111] & -11.7 & -0.0267 [-0.0389, -0.0146] & -3.5 & 0.022 & -0.078 & 12 \\
A & C2 & 0.8 & -0.0093 [-0.0160, -0.0025] & -5.4 & -0.0148 [-0.0249, -0.0048] & -1.9 & 0.031 & -0.040 & 12 \\
A & C2 & 0.9 & -0.0027 [-0.0074, 0.0021] & -1.4 & -0.0045 [-0.0126, 0.0036] & -0.6 & 0.032 & -0.020 & 12 \\
A & C3 & 0.4 & -0.0897 [-0.0960, -0.0833] & -53.5 & -0.1705 [-0.1880, -0.1530] & -22.6 & -0.343 & -0.469 & 12 \\
A & C3 & 0.6 & -0.0459 [-0.0577, -0.0341] & -26.9 & -0.0693 [-0.0877, -0.0509] & -9.2 & -0.087 & -0.207 & 12 \\
A & C3 & 0.8 & -0.0104 [-0.0172, -0.0036] & -6.1 & -0.0184 [-0.0280, -0.0089] & -2.4 & 0.001 & -0.041 & 12 \\
A & C3 & 0.9 & -0.0071 [-0.0116, -0.0026] & -4.2 & -0.0095 [-0.0153, -0.0037] & -1.3 & 0.038 & -0.032 & 12 \\
B & C1 & 0.4 & 0.0003 [-0.0049, 0.0055] & 0.2 & -0.0024 [-0.0059, 0.0011] & -0.3 & -0.325 & 0.020 & 12 \\
B & C1 & 0.6 & 0.0026 [-0.0010, 0.0062] & 1.4 & -0.0023 [-0.0074, 0.0028] & -0.3 & -0.074 & -0.001 & 12 \\
B & C1 & 0.8 & 0.0023 [-0.0010, 0.0057] & 1.2 & 0.0000 [-0.0033, 0.0033] & 0.0 & -0.031 & 0.007 & 12 \\
B & C1 & 0.9 & 0.0021 [-0.0005, 0.0046] & 1.1 & -0.0000 [-0.0021, 0.0021] & -0.0 & -0.001 & 0.002 & 12 \\
B & C2 & 0.4 & 0.0008 [-0.0030, 0.0046] & 0.4 & -0.0015 [-0.0062, 0.0032] & -0.2 & -0.137 & 0.039 & 12 \\
B & C2 & 0.6 & 0.0023 [-0.0008, 0.0054] & 1.3 & -0.0012 [-0.0033, 0.0009] & -0.1 & -0.060 & 0.013 & 12 \\
B & C2 & 0.8 & 0.0020 [-0.0006, 0.0047] & 1.1 & -0.0002 [-0.0025, 0.0021] & -0.0 & 0.016 & -0.001 & 12 \\
B & C2 & 0.9 & 0.0041 [0.0009, 0.0072] & 2.1 & 0.0004 [-0.0017, 0.0025] & 0.0 & 0.015 & 0.002 & 12 \\
B & C3 & 0.4 & -0.0067 [-0.0112, -0.0023] & -3.4 & -0.0111 [-0.0166, -0.0055] & -1.4 & -0.589 & 0.011 & 12 \\
B & C3 & 0.6 & -0.0036 [-0.0090, 0.0018] & -1.8 & -0.0051 [-0.0093, -0.0009] & -0.6 & -0.204 & 0.001 & 12 \\
B & C3 & 0.8 & 0.0017 [-0.0023, 0.0056] & 0.9 & -0.0016 [-0.0051, 0.0019] & -0.2 & -0.017 & 0.012 & 12 \\
B & C3 & 0.9 & -0.0002 [-0.0034, 0.0031] & 0.0 & -0.0008 [-0.0029, 0.0013] & -0.1 & -0.049 & 0.010 & 12 \\
\end{longtable}
\end{center}
\end{landscape}

\begin{landscape}
\begin{center}
\scriptsize
\setlength{\tabcolsep}{3pt}
\renewcommand{\arraystretch}{1.05}
\begin{longtable}{llr P{4.2cm} r P{4.2cm} r r r r}
\caption{Full $s$ sweep results for feature set F1 (primitives plus engineered ratio). Same conventions as Table~\ref{tab:app_v3_f0}.}\label{tab:app_v3_f1}\\
\toprule
DGP & Arm & $s$ & $\Delta$PR-AUC (95\% CI) & rel$\Delta$PR (\%) & $\Delta$ROC-AUC (95\% CI) & rel$\Delta$ROC (\%) & $\Delta$\texttt{cooc\_mean} & $\Delta$\texttt{latent\_corr} & $n$ \\
\midrule
\endfirsthead
\multicolumn{10}{l}{\textit{Table \thetable\ continued from previous page}}\\
\toprule
DGP & Arm & $s$ & $\Delta$PR-AUC (95\% CI) & rel$\Delta$PR (\%) & $\Delta$ROC-AUC (95\% CI) & rel$\Delta$ROC (\%) & $\Delta$\texttt{cooc\_mean} & $\Delta$\texttt{latent\_corr} & $n$ \\
\midrule
\endhead
\bottomrule
\endfoot
\bottomrule
\endlastfoot
A & C1 & 0.4 & 0.0008 [-0.0053, 0.0069] & 0.6 & -0.0056 [-0.0111, -0.0002] & -0.7 & 0.046 & 0.004 & 12 \\
A & C1 & 0.6 & 0.0009 [-0.0053, 0.0070] & 0.5 & -0.0003 [-0.0027, 0.0022] & -0.0 & 0.024 & 0.019 & 12 \\
A & C1 & 0.8 & 0.0036 [0.0004, 0.0068] & 1.6 & 0.0002 [-0.0013, 0.0018] & 0.0 & 0.006 & 0.012 & 12 \\
A & C1 & 0.9 & 0.0017 [-0.0008, 0.0042] & 0.7 & -0.0001 [-0.0015, 0.0014] & -0.0 & 0.003 & -0.001 & 12 \\
A & C2 & 0.4 & 0.0012 [-0.0028, 0.0051] & 0.5 & -0.0075 [-0.0140, -0.0010] & -0.9 & 0.075 & -0.020 & 12 \\
A & C2 & 0.6 & 0.0034 [-0.0001, 0.0068] & 1.4 & -0.0008 [-0.0030, 0.0013] & -0.1 & 0.048 & 0.011 & 12 \\
A & C2 & 0.8 & 0.0002 [-0.0027, 0.0031] & 0.2 & 0.0009 [-0.0007, 0.0025] & 0.1 & 0.014 & 0.010 & 12 \\
A & C2 & 0.9 & 0.0018 [-0.0004, 0.0040] & 0.8 & 0.0004 [-0.0009, 0.0016] & 0.0 & 0.007 & 0.028 & 12 \\
A & C3 & 0.4 & -0.0085 [-0.0123, -0.0048] & -3.5 & -0.0118 [-0.0187, -0.0050] & -1.5 & 0.064 & 0.020 & 12 \\
A & C3 & 0.6 & -0.0020 [-0.0076, 0.0036] & -0.8 & -0.0062 [-0.0111, -0.0013] & -0.8 & 0.086 & -0.003 & 12 \\
A & C3 & 0.8 & 0.0027 [-0.0002, 0.0057] & 1.2 & -0.0019 [-0.0042, 0.0004] & -0.2 & 0.045 & -0.011 & 12 \\
A & C3 & 0.9 & 0.0013 [-0.0010, 0.0036] & 0.6 & -0.0008 [-0.0028, 0.0012] & -0.1 & 0.016 & 0.005 & 12 \\
B & C1 & 0.4 & 0.0051 [-0.0002, 0.0105] & 2.7 & 0.0010 [-0.0012, 0.0031] & 0.1 & 0.245 & 0.002 & 12 \\
B & C1 & 0.6 & 0.0048 [-0.0010, 0.0107] & 2.5 & 0.0017 [0.0000, 0.0034] & 0.2 & 0.298 & -0.014 & 12 \\
B & C1 & 0.8 & 0.0019 [-0.0038, 0.0077] & 1.0 & 0.0008 [-0.0011, 0.0026] & 0.1 & 0.135 & -0.003 & 12 \\
B & C1 & 0.9 & 0.0022 [-0.0033, 0.0077] & 1.2 & 0.0009 [-0.0003, 0.0021] & 0.1 & 0.075 & -0.008 & 12 \\
B & C2 & 0.4 & 0.0036 [-0.0020, 0.0092] & 1.8 & 0.0005 [-0.0014, 0.0024] & 0.1 & 0.469 & -0.021 & 12 \\
B & C2 & 0.6 & 0.0032 [-0.0033, 0.0096] & 1.6 & 0.0008 [-0.0017, 0.0034] & 0.1 & 0.394 & -0.018 & 12 \\
B & C2 & 0.8 & 0.0018 [-0.0036, 0.0071] & 0.9 & 0.0010 [-0.0012, 0.0031] & 0.1 & 0.167 & -0.005 & 12 \\
B & C2 & 0.9 & 0.0011 [-0.0041, 0.0062] & 0.5 & -0.0002 [-0.0025, 0.0021] & -0.0 & 0.070 & 0.005 & 12 \\
B & C3 & 0.4 & -0.0030 [-0.0073, 0.0012] & -1.5 & -0.0083 [-0.0137, -0.0028] & -1.0 & 0.240 & -0.014 & 12 \\
B & C3 & 0.6 & 0.0027 [-0.0022, 0.0076] & 1.4 & -0.0009 [-0.0040, 0.0022] & -0.1 & 0.405 & -0.008 & 12 \\
B & C3 & 0.8 & 0.0058 [-0.0008, 0.0124] & 3.0 & 0.0020 [0.0002, 0.0038] & 0.3 & 0.296 & -0.000 & 12 \\
B & C3 & 0.9 & 0.0041 [-0.0013, 0.0095] & 2.1 & 0.0008 [-0.0007, 0.0024] & 0.1 & 0.155 & -0.012 & 12 \\
\end{longtable}
\end{center}
\end{landscape}

\section{Formal definitions of the data generating processes}

\subsection{DGP-A}

Latents:
\begin{align}
U &\sim \mathcal{N}(0,\sigma_U^2), \\
V &\sim \mathcal{N}(0,\sigma_V^2).
\end{align}
Primitives:
\begin{align}
\log a &= U + V + \epsilon_a, \\
\log b &= U - V + \epsilon_b, \\
\epsilon_a, \epsilon_b &\sim \mathcal{N}(0,\sigma_\epsilon^2).
\end{align}
Label:
\begin{align}
y &\sim \text{Bernoulli}\left(\sigma(\beta V)\right).
\end{align}
Control feature:
\begin{align}
r &= \log a - \log b.
\end{align}

\subsection{DGP-B}

Exposure:
\begin{align}
\log E \sim \mathcal{N}(\mu_E, \sigma_E^2).
\end{align}
Signal:
\begin{align}
V \sim \mathcal{N}(0,\sigma_V^2).
\end{align}
Counts:
\begin{align}
A \mid E,V &\sim \text{NegBin}(\lambda_A, \phi), \quad \lambda_A = E \cdot \exp(V), \\
B \mid E,V &\sim \text{NegBin}(\lambda_B, \phi), \quad \lambda_B = E \cdot \exp(-V).
\end{align}
Label:
\begin{align}
y &\sim \text{Bernoulli}\left(\sigma(\beta V)\right).
\end{align}
Control feature:
\begin{align}
r &= \log(A+s_0) - \log(B+s_0).
\end{align}

\section{Definition of the path co-usage metric}

For a given tree and a given leaf, consider the set of split features on the path from the root to that leaf.
Define an indicator that equals one if both target primitives appear on that path.
Weight each leaf by its cover.
Average across leaves and trees.
This yields \texttt{cooc\_path\_mean}.

This metric is structural.
It is not a proof of internal computation.

\section{Confidence intervals}
\label{app:ci}

Let $d_i$ be the paired delta for a fixed setting, computed against the matched C0 baseline.
The reported estimate is the sample mean $\bar d = \frac{1}{n}\sum_{i=1}^n d_i$.
Let $s_d$ be the sample standard deviation of the $d_i$ with \texttt{ddof}=1.
The standard error is $\mathrm{SE}=s_d/\sqrt{n}$.
The 95\% interval is $\bar d \pm 1.96\cdot \mathrm{SE}$.

\section{Additional sweeps and plots}

\begin{figure}[t]
\centering
\includegraphics[width=\textwidth]{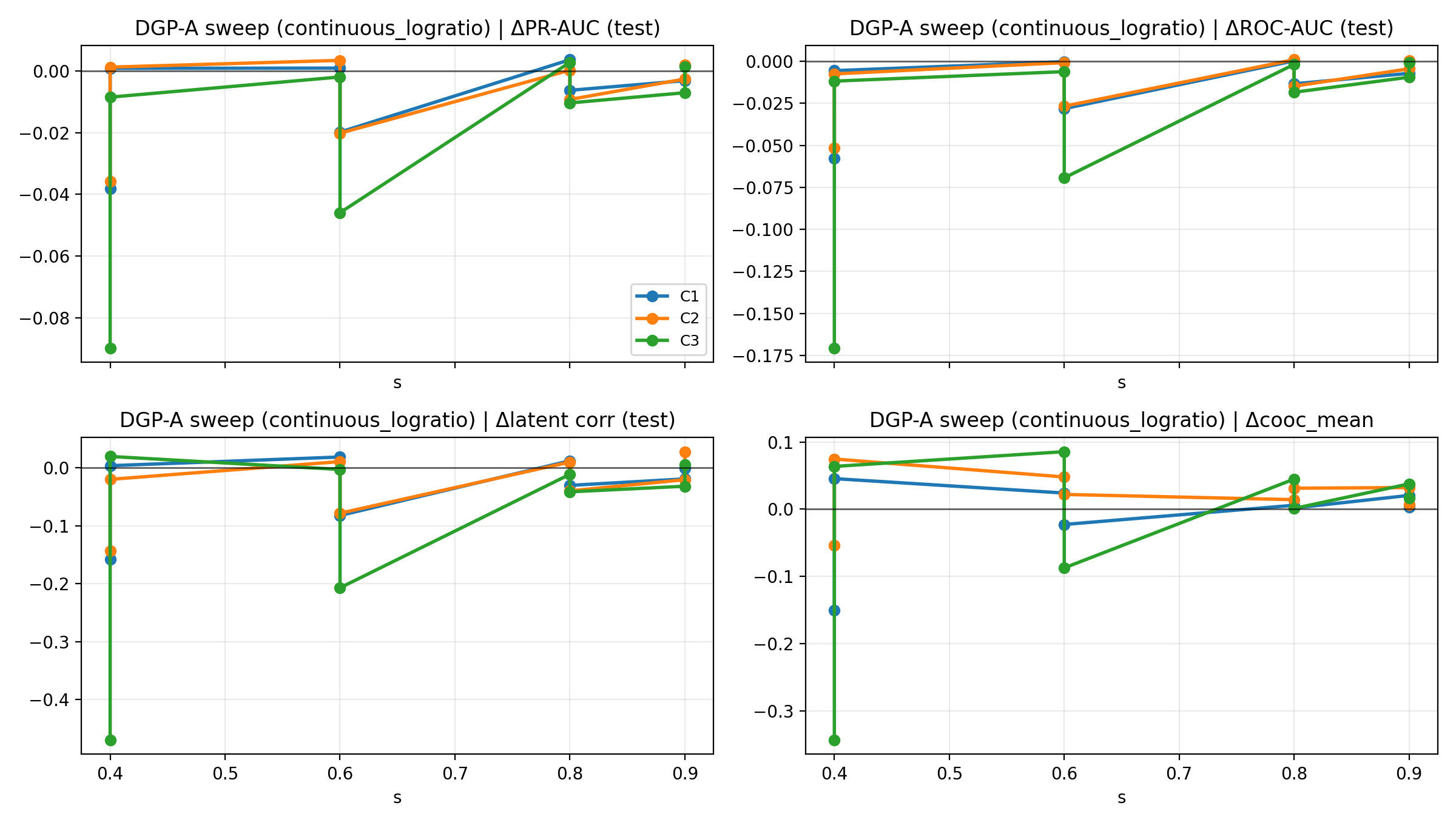}
\caption{DGP-A low $s$ sweep. The sweep includes $s \in \{0.4,0.6,0.8,0.9\}$. The interaction synthesis setting F0 is more sensitive to stronger intra-tree subsampling.}
\label{fig:low_s_A}
\end{figure}

\begin{figure}[t]
\centering
\includegraphics[width=\textwidth]{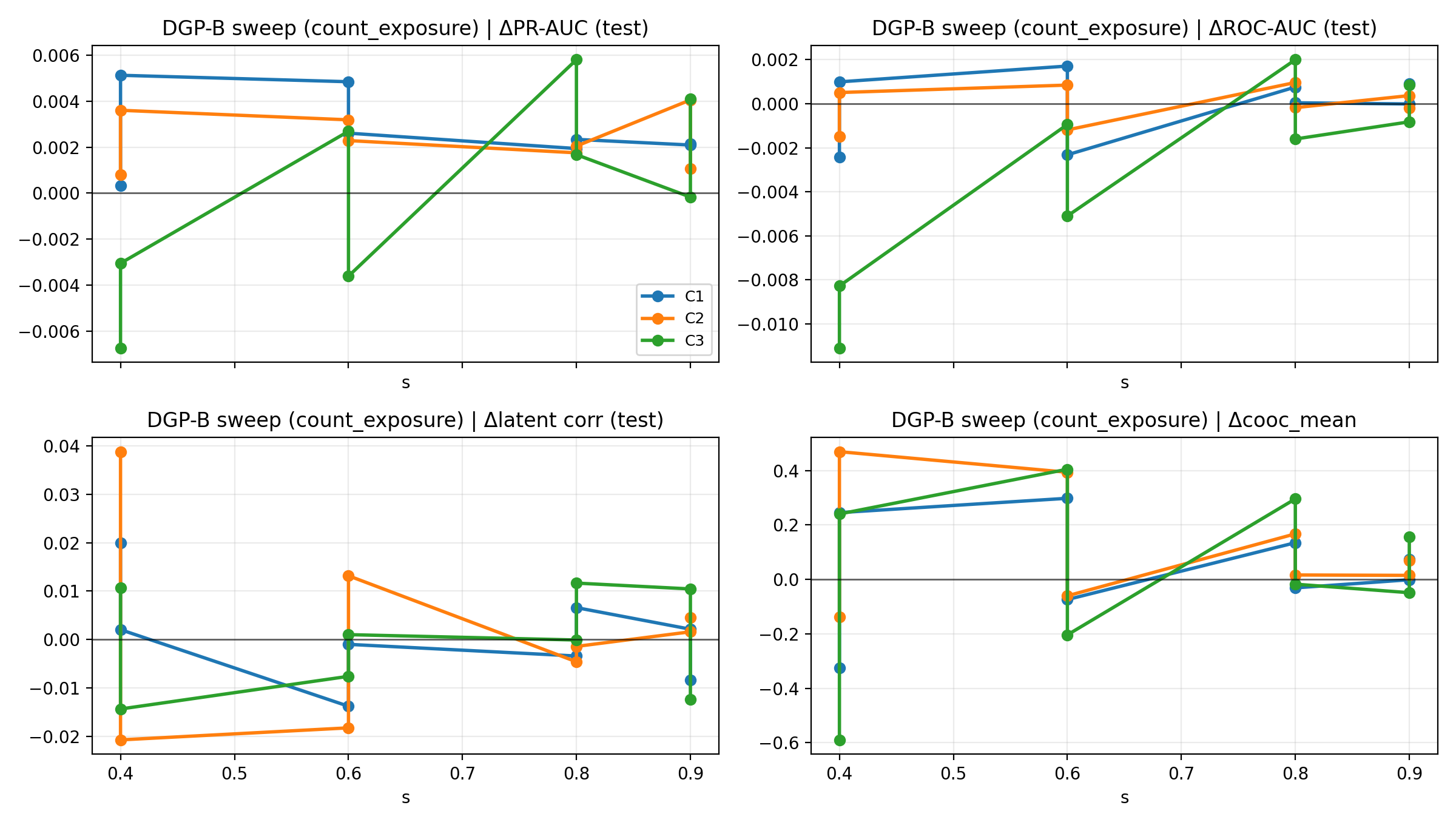}
\caption{DGP-B low $s$ sweep. The same qualitative pattern holds under the count plus exposure construction.}
\label{fig:low_s_B}
\end{figure}

\begin{figure}[t]
\centering
\includegraphics[width=\textwidth]{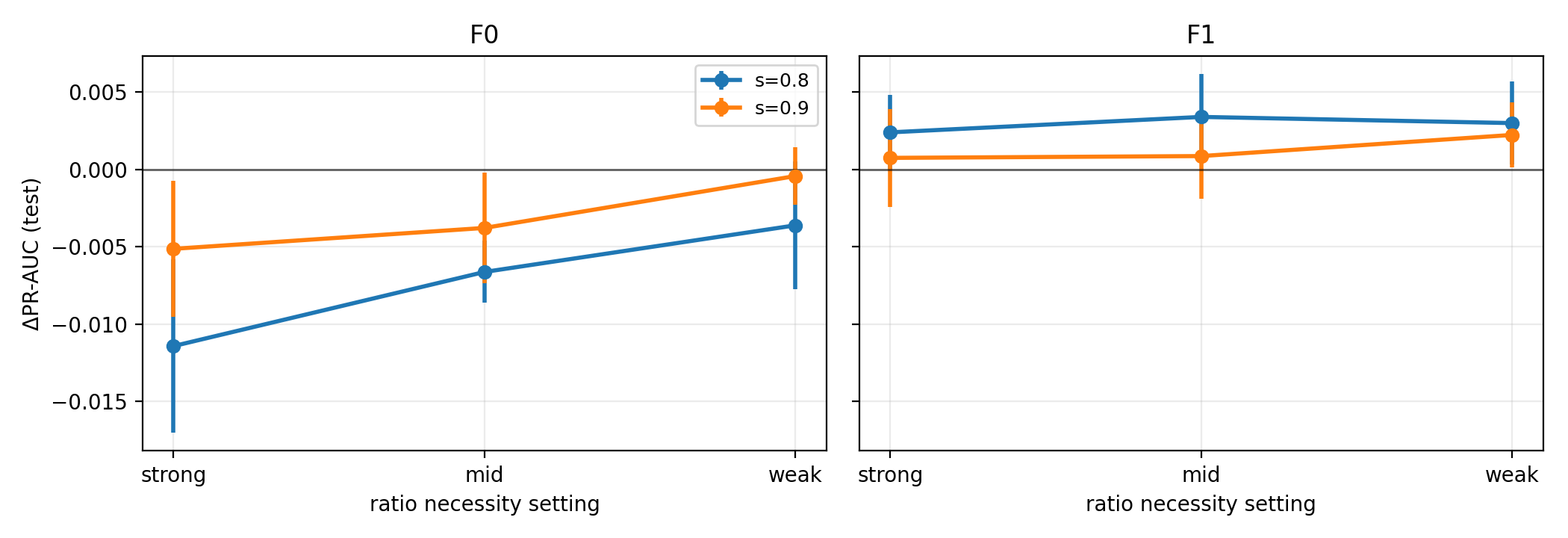}
\caption{DGP-A boundary sweep under C3. The ratio necessity setting varies the strength of nuisance cancellation. The drop in F0 is largest when the ratio is most necessary.}
\label{fig:boundary}
\end{figure}

\end{document}